\documentclass[acmsmall,scree]{acmart}

\AtBeginDocument{%
  \providecommand\BibTeX{{%
    \normalfont B\kern-0.5em{\scshape i\kern-0.25em b}\kern-0.8em\TeX}}}

\usepackage{multirow}
\usepackage{multicol}
\usepackage{listings}
\usepackage{makecell}
\usepackage{xcolor}

\newcommand{\revise}[1]{\textcolor{black}{#1}}

\begin{document}

\title{CodeKGC: Code Language Model for Generative \\ Knowledge Graph Construction}




\author{Zhen Bi}
\email{bizhen_zju@zju.edu.cn}
\authornotemark[1]
\author{Jing Chen}
\authornote{Both authors contributed equally to this research.}
\email{jingc0116@gmail.com}
\affiliation{%
  \institution{Zhejiang University}
  \city{Hangzhou}
  \state{Zhejiang}
  \country{China}
}
\affiliation{%
  \institution{Zhejiang University - Ant Group Joint Laboratory of Knowledge Graph}
  \city{Hangzhou}
  \state{Zhejiang}
  \country{China}
}

\author{Yinuo Jiang}
\affiliation{%
  \institution{Zhejiang University}
  \city{Hangzhou}
  \state{Zhejiang}
  \country{China}
}
\affiliation{%
  \institution{Zhejiang University - Ant Group Joint Laboratory of Knowledge Graph}
  \city{Hangzhou}
  \state{Zhejiang}
  \country{China}
}
\email{3200100732@zju.edu.cn}

\author{Feiyu Xiong}
\email{feiyu.xfy@zju.edu.cn}
\author{Wei Guo}
\email{huaisu@taobao.com}
\affiliation{
  \institution{Alibaba Group}
  \city{Hangzhou}
  \state{Zhejiang}
  \country{China}
}





\author{Huajun Chen}
\email{huajunsir@zju.edu.cn}
\author{Ningyu Zhang}
\authornote{Corresponding Author.}
\email{zhangningyu@zju.edu.cn}
\affiliation{%
  \institution{Zhejiang University}
  \city{Hangzhou}
  \state{Zhejiang}
  \country{China}
}
\affiliation{%
  \institution{Zhejiang University - Ant Group Joint Laboratory of Knowledge Graph}
  \city{Hangzhou}
  \state{Zhejiang}
  \country{China}
}


\newcommand{\ours}{\textbf{CodeKGC}}

\begin{abstract}
Current generative knowledge graph construction approaches usually fail to capture structural knowledge by simply flattening natural language into serialized texts or a specification language. However, large generative language model trained on structured data such as code has demonstrated impressive capability in understanding natural language for structural prediction and reasoning tasks. Intuitively, we address the task of generative knowledge graph construction with code language model: given a code-format natural language input, the target is to generate triples which can be represented as code completion tasks. Specifically, we develop schema-aware prompts that effectively utilize the semantic structure within the knowledge graph. As code inherently possesses structure, such as class and function definitions, it serves as a useful model for prior semantic structural knowledge. Furthermore, we employ a rationale-enhanced generation method to boost the performance. Rationales provide intermediate steps, thereby improving knowledge extraction abilities. Experimental results indicate that the proposed approach can obtain better performance on benchmark datasets compared with baselines\footnote{Code and datasets are available in \url{https://github.com/zjunlp/DeepKE/tree/main/example/llm}}.
\end{abstract}




\maketitle

\section{Introduction}
\label{sec:introduction}

Knowledge Graphs (KGs) typically consist of a collection of nodes (entities) and edges (relationships) that interconnect these entities. 
High-quality KGs predominantly depend on human-curated structured or semi-structured data \cite{KGC-survey}.
Consequently, Knowledge Graph Construction (KGC) has been proposed as a method for populating a KG with new knowledge components, such as entities and relations.
Existing techniques such as \cite{CasRel,DBLP:conf/naacl/ZhongC21}, have achieved considerable success by employing pipelines that systematically integrate named entity recognition \cite{DBLP:journals/tkde/LiSHL22}, entity linking \cite{DBLP:conf/emnlp/WuPJRZ20}, and relation extraction \cite{DBLP:conf/naacl/ZhangDSWCZC19}.
Nonetheless, these approaches introduce certain limitations, including the propagation of errors and overlapping issues, which can adversely affect the quality and accuracy for the  knowledge graph construction.

To address the accumulation of errors and overlapping issues, it is often recommended to design end-to-end approaches.
Framing knowledge graph construction as end-to-end natural language generation is an increasingly promising technique that has led to empirical success \cite{DBLP:journals/corr/abs-2210-14698}.
This approach involves converting relational structures into serialized texts and using expressive black-box neural networks, such as pre-trained language models (LMs) \cite{DBLP:conf/acl/LewisLGGMLSZ20,t5}, to predict flattened texts that encode the relational structure.
Generative language models have demonstrated their effectiveness in various knowledge graph construction tasks such as entity extraction \cite{DBLP:conf/acl/YanGDGZQ20}, entity typing \cite{GET}, relation extraction \cite{GenIE}, triple extraction \cite{CGT}, and event extraction \cite{Text2Event}.

\begin{figure}
\includegraphics[width=1.0\textwidth]{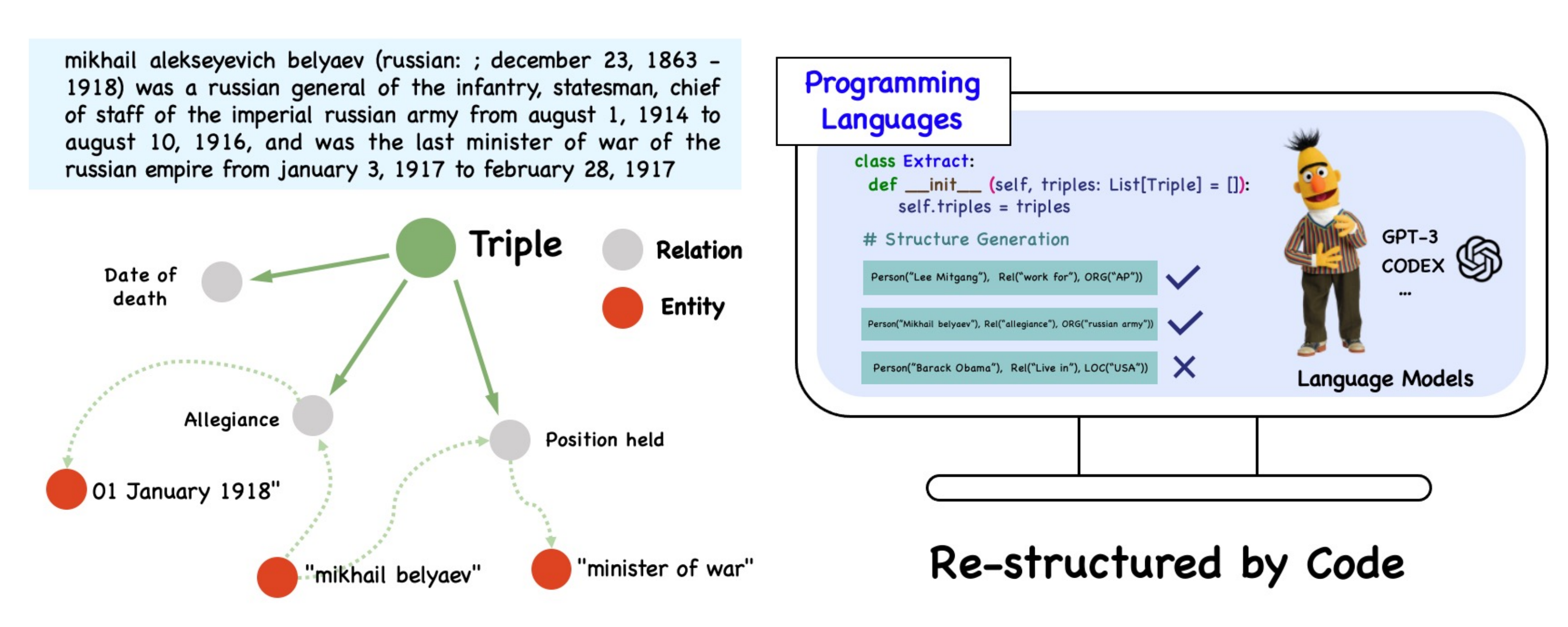}
\caption{
Code language model for generative knowledge graph construction.
The structure of code can help to understand complex structural information in natural language and we design structural code prompts as input into the language models.}
\label{figure:intro}
\end{figure}

Despite the empirical success, it is impractical to expect neural networks can easily handle intricate intra-structure dependencies, especially for scenarios where facts are complexly overlapped.
Language models struggle to generate these unnatural outputs because the syntax and structure of this type of text is different from the free-form text in the training data.
Furthermore, generating structured sentences that push neighboring nodes (overlapping facts) farther apart is difficult since semantically relevant words often occur within close proximity to each other.
Take Figure \ref{figure:intro} as an instance, 
the relations between two entities can stretch across many words, and flattening such structured information into a string may result in poor generation performance.
Recent progress in large-language models of \emph{code} (Code-LLMs) \cite{Codex} has demonstrated the potential to employ Code-LLMs for complex reasoning \cite{POT},  structured commonsense reasoning \cite{COCOGEN} and
\revise{structure prediction \cite{Code4Struct} tasks}.
Our main insight is that \emph{language models of code may model structure better than those of natural language}. 
Intuitively, we formulate knowledge graph construction  as code generation tasks, which can augment pre-trained language models through explicit modeling of the structure.

Therefore, we propose a schema-aware \textbf{Code} prompt and rationale-enhanced \textbf{K}nowledge \textbf{G}raph \textbf{C}onstruction method dubbed {\ours}.
Inspired by the great structural and reasoning abilities of Code-LLMs, we convert the natural language to the code formats instead of representing them as fat word strings.
We design structural code prompts that encode schema information to maintain intrinsic semantic structures.
By utilizing base class definitions, we can effectively model pre-defined entity and relation schema in the KGs.
We also propose a rationale-enhanced generation method for generating entities and relations within the model, which helps to improve the ability of knowledge extraction.
During the generation procedure, a rational-enhanced module is leveraged to elicit the power of language models.
To demonstrate the effectiveness of our approach, we apply {\ours} to three datasets: ADE, CONLL04 and SciERC.
We evaluate the performance of our proposed model in both zero-shot and few-shot settings.
Experimental results on benchmark datasets show that the proposed approach can obtain better performance compared with baselines.
These results demonstrate the effectiveness of the proposed method, and its potential to solve complex language modeling problems.

\subsection{Contribution}
The contributions of our study can be summarized as follows: 
\begin{itemize}
\item
\revise{
First, we propose \ours, a novel method which re-structures natural language into code formats for generative knowledge graph construction.
Our proposed approach is the first to re-structure code data for the knowledge graph construction task.
}

\item 
\revise{
Second, we design effective code prompting approach to encode the schema in the KGs through the effective modeling of code definitions. 
Particularly,
we also propose a novel rationale-enhanced generation method to improve the  ability of knowledge extraction.
Our designed code-formatted approach enables the language model to efficiently extract the requisite entities and relations from the text.
}

\item Finally, we conduct extensive experiments to evaluate the performance of the proposed model.
We evaluate our proposed method on three datasets and demonstrate that \revise{\ours~ outperforms all baseline models in various evaluation settings, particularly in 
addressing complex structure extraction problems such as overlapping issues.}

\end{itemize}

\subsection{Outline of the Article}
The remainder of this article is structured as follows. 
In Section \ref{sec:related_work}, we introduce related works on knowledge graph construction and provide a detailed description of the code large language model and re-structured learning.
In Section \ref{sec:method}, we first give an overview of our designed {\ours}, then explain the schema-aware code prompt design and rational-enhanced generation. 
In Section \ref{sec:experiment}, we present the results of experiments on three knowledge graph construction datasets, which demonstrate the effectiveness of our proposed method. 
Additionally, we provide an ablation study to analyze our proposed model.
In Section \ref{sec:discussion}, we discuss the potential reasons for the potential of code language models in structural and reasoning tasks.
In Section \ref{sec:conclusion}, we conclude the article with discussions on future works.

\section{Related work}
\label{sec:related_work}

\subsection{Knowledge Graph Construction}
The knowledge graph construction task mainly focus on extracting structural information from unstructured texts.
Specifically, extracting relational triples from the text plays a vital role in constructing large-scale knowledge graphs.
\citet{CasRel} revisit the relational triple extraction task and propose a novel cascade binary tagging framework (CasRel) derived from a principled problem formulation.
\citet{BiRTE} propose a bidirectional extraction framework-based method that extracts triples based on the entity pairs extracted from two complementary directions.
For zero-shot relation triplet extraction,  \citet{RelationPrompt} proposes to synthesize relation examples by prompting language models to generate structured texts.
\citet{UIE} propose a unified text-to-structure generation framework, namely UIE, which can universally model different IE tasks.
\citet{CasRel} reexamines the task of relational triple extraction and introduces a novel cascading binary tagging framework (CasRel) originating from a well-founded problem formulation.
\citet{BiRTE} proposes a method based on a bidirectional extraction framework that extracts triples by leveraging entity pairs obtained from two complementary directions.
For zero-shot relation triplet extraction, \citet{RelationPrompt} suggests synthesizing relation examples by prompting pre-trained language models.
\citet{UIE} introduces a unified text-to-structure generation framework called UIE, capable of universally modeling various IE tasks.
Nevertheless, most mainstream methods take the serialized sequence as the input and ignore the natural structure format of  language and semantic knowledge.
To the best of our knowledge, our proposed approach is the first to re-structure code data for the knowledge graph construction task.

\subsection{Code Large Language Model}
Code language models \cite{Unixcode,CodeBERT, CodeT5}, have showcased their extraordinary prowess in generating code and can now tackle basic programming tasks.
\citet{Codex} represents a GPT language model optimized using publicly accessible GitHub code, with research focusing on its ability to write Python code.
\citet{AlphaCode} introduces a system devised for code generation that can generate innovative solutions to challenges demanding more sophisticated reasoning.
Recently, the combination of language models and code, facilitated by the emergence of code language models, has proven to be effective in tackling specific tasks \cite{reasoning-survey}.
\citet{COCOGEN} defines structured commonsense reasoning tasks as code generation tasks, demonstrating that pre-trained code-based LMs exhibit superior structured commonsense reasoning capabilities compared to natural language LMs.
\revise{
\citet{Code4Struct} harnesses the text-to-structure translation proficiency to address structured prediction tasks within the NLP domain.
}
\citet{PAL} introduces a novel technique known as PAL, which employs the large language model to interpret natural language problems and create programs as intermediate reasoning steps. However, the solution phase is delegated to a programmatic runtime, such as a Python interpreter.
\citet{POT} proposes program of thoughts (PoT) which disentangles computation from reasoning.
However, for the task of knowledge graph construction, how to utilize code language to model structural information has not been studied.

\subsection{Re-structured Learning}
In the re-structured learning paradigm \citep{reStructured}, the importance of data is emphasized, with model pre-training and fine-tuning for downstream tasks considered as processes of data storage and retrieval. \citet{DeepStruct} presents a method to enhance the structural comprehension abilities of language models by pre-training them on task-agnostic corpora to generate structures from text. \citet{TANL} proposes a novel framework for addressing various structured prediction language tasks, such as joint entity and relation extraction, nested named entity recognition, and more.
\citet{DEEPEX} illustrates that a straightforward pre-training task of predicting the correspondence between relational information and input text is an effective approach to generate task-specific output. The transformation of natural language into structured language has also proven to be efficient for universal information extraction \citep{UIE}.
Our work is distinct from previous re-structured learning approaches in that we leverage the naturally structured and semantically-rich code language to enhance the learning of structural prompts.

\section{CodeKGC: Code Language Model for Generative Knowledge Graph Construction}
\label{sec:method}

We introduce a novel method, named {\ours},  which re-structures natural language into code language for generative knowledge graph construction.
As shown in Figure \ref{figure:model}, {\ours} consists of two main components: schema-aware prompt construction (Section \ref{method:schema}), the rationale-enhanced generation (Section \ref{method:rationale}).

\subsection{Schema-aware Prompt Construction}
\label{method:schema}

We frame the knowledge graph construction task as the code generation problem.
Each training instance for such tasks is in the form $(\mathcal{T},\mathcal{G})$,
where $\mathcal{T}$ is a text input, and $\mathcal{G}$ is the knowledge graph to be generated.
It should be noted that $\mathcal{G}$ are triples which consists of entities and their connected relations.
Since LMs trained on free-form natural language text are likely to fail to capture the topology of the relational structure, as shown in Figure \ref{figure:intro}, the key idea of our method is to transform a set of output triples (graph $\mathcal{G}$) into a semantically equivalent program language written in Python (graph $\mathcal{G}_c$).
We use the pre-defined script, which conforms to the grammatical structure of Python, to transform the original sample $(\mathcal{T},\mathcal{G})$ into code format $(\mathcal{T}_c,\mathcal{G}_c)$.
The re-structured language will be fed into the code language model by our proposed {\ours}.
When doing the inference, we eventually convert it back into the graph $\mathcal{G}$.

\begin{figure*}
\centering
\includegraphics[width=0.95\textwidth]{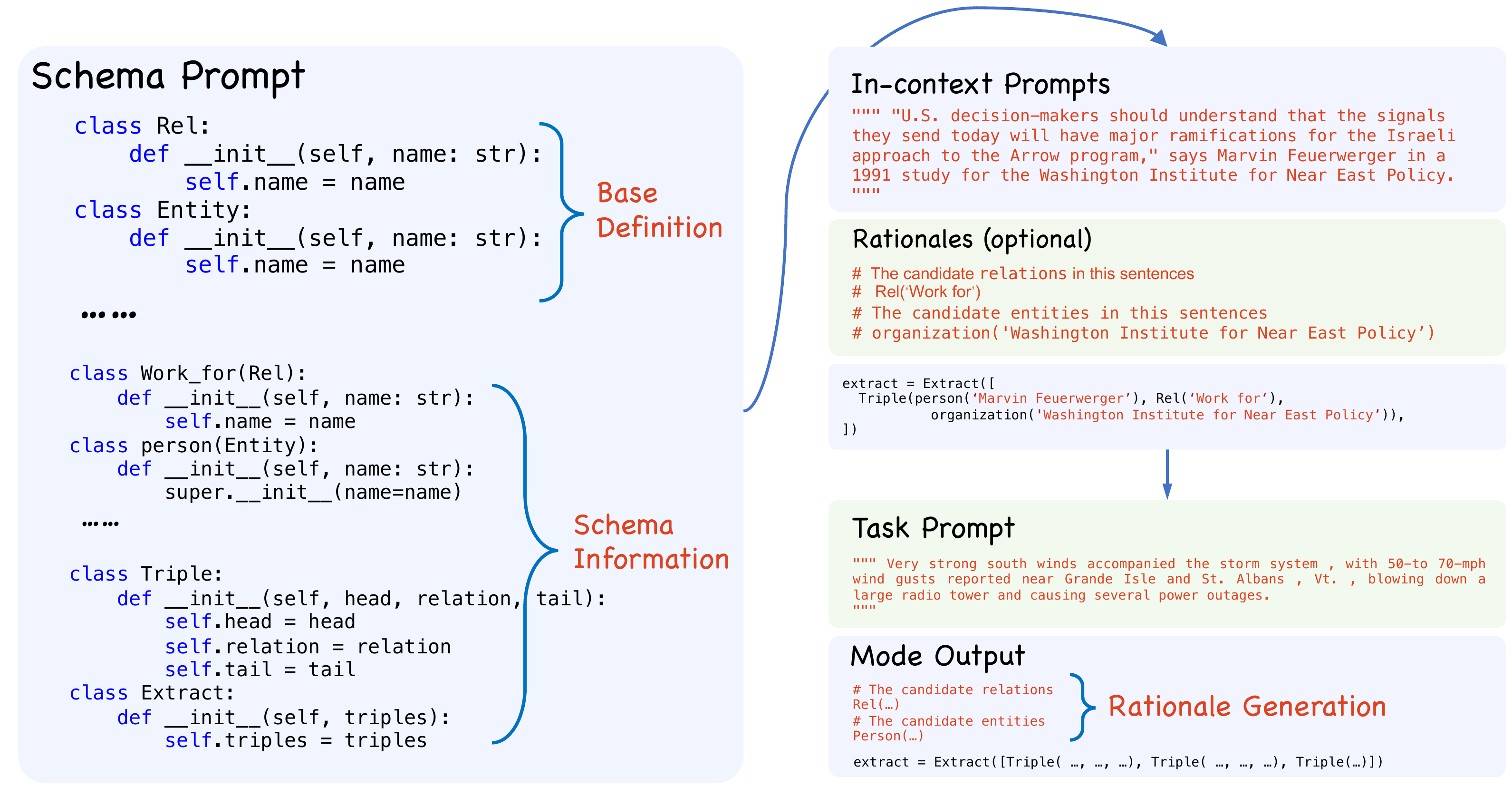}
\caption{
\revise{
Overview of the proposed \ours~ with code languages. 
}
The original natural language is converted into code formats and then fed into the code language model which is guided by a specified task prompt.
We use schema-aware prompt to preserve the relations, properties, and constraints in the knowledge graph.
Inspired by \cite{COT}, {\ours} also utilizes an optional rationale-enhanced module as an intermediate reasoning step in the in-context learning samples.
}
\label{figure:model}
\end{figure*}

Specifically, we convert the sentence pair into a series of pre-defined classes, and construct a task prompt containing a \texttt{Docstrings} which describes the task.
An example of a task prompt can be found in the bottom right of Figure \ref{figure:model}.
We define a totally new class \texttt{Extract} to do the extraction process from the \texttt{Docstrings}.
The incomplete task prompt we build here is used to guild the output of the code language model.
With the re-structured code formats which retain the syntax and structural features, the code language model tends to generate more accurate relations and entities.

The schema in the KG describes the structure of the data, including the entities, relations, properties and constraints that make up the graph.
We can integrate the schema information by the inheritance of Python class definitions. 
As shown in Figure \ref{figure:model}, we define base classes  \texttt{Entity} and \texttt{Relation} that describe the general definitions for relations and entities for knowledge graph construction.
For example, to represent the entities that belong to the type "person", the \texttt{Person} class should inherit from the base \texttt{Entity} class.
We further define \texttt{Triple} class to represent the triples contained in the textual input.
\texttt{Triple} class consists of the head-tail entity pairs and their corresponding relationship.
Each triple will be represented as the instance of their corresponding entity and relation type classes.
For example, if there exists a triplet $(London, located in, UK)$, we represent it with our designed code formats, \texttt{Triple}(\texttt{LOC}("London"),  \texttt{Rel}("located in"),  \texttt{LOC}("London")).
It should be noted that the input augments for task prompt \texttt{Extract} include a list of  \texttt{Triple} instances.
By this way, we can naturally model complex structural information in language, such as nested entities or relationships.
Even for the constraints between entity types, we can define specific instantiated triple classes to model this information.

\begin{figure*}
\centering
\includegraphics[width=0.5\textwidth]{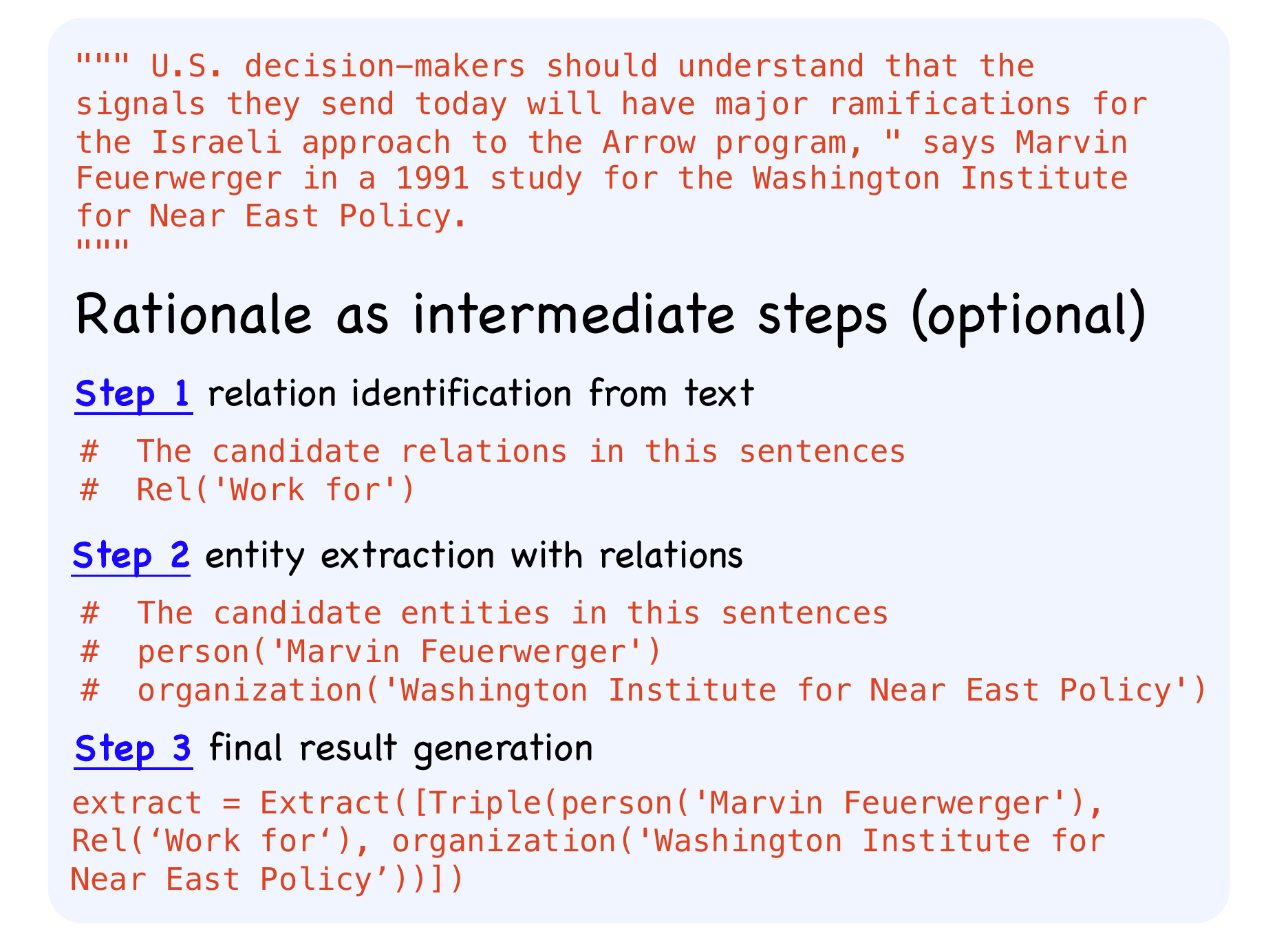}
\caption{
\revise{Code generation with rationales.
}
We try to decompose the  task of knowledge graph constructions into multiple intermediate steps.
By leveraging logical reasoning chains like chain-of-thought \cite{COT}, {\ours} specializes in solving complex structural patterns.
}
\label{figure:rationale}
\end{figure*}

\subsection{Rationale-enhanced Generation}
\label{method:rationale}
To perform the task on an unseen example, we give the code LM a few exemplars of completed ground-truth relational triples that are composed of task prompts.
We select $k$ samples from the dataset that include all $n$ relation types and concatenate them with the task prompts for $n$-shot learning.
Note that the general ability to discover new content, such as class definitions of new entities and relations, can help to generate new instances to boost the KG construction performance.

Inspired by chain of thoughts \cite{COT}, we propose an optional rationale-enhanced generation method to improve the underlying reasoning abilities. 
For normal prompting, given the textual code input $\mathcal{T}_c$,
prompt $\mathcal{P}$, 
and the and probabilistic model $p_{LM}$, we aim to maximize the likelihood of the  $\mathcal{G}_c$ as:

\begin{equation}
    p(\mathcal{G}_c|\mathcal{T}_c, \mathcal{P}) = \prod_{i}^{|\mathcal{G}_c|} p_{LM}(g_i, | \mathcal{T}_c, \mathcal{P}, g_{<i} )
    \label{equation:1}
\end{equation}

where $g_i$ is the $i$-th token of $\mathcal{G}_c$, and $|\mathcal{G}_c|$ denotes the length of $\mathcal{G}_c$ which will be generated.
For few-shot prompting, $\mathcal{P}$ is comprised of $k$ exemplars of $(\mathcal{T}_c,\mathcal{G}_c)$ pairs.
We have designed a specific rationale to enhance the reasoning capabilities of language models.
Therefore, we add rationales $\mathcal{R}$ (reasoning steps)  into prompt where $\mathcal{P}=\{ (\mathcal{T}_c^{i}, \mathcal{R}^{i}, \mathcal{G}_c^{i}) \}_{i=1}^{k} $ , thus Equation \ref{equation:1} can be reformed to:

\begin{equation}
    p(\mathcal{G}_c|\mathcal{T}_c, \mathcal{P}) =     p(\mathcal{G}_c|\mathcal{T}_c, \mathcal{P}, \mathcal{R})
    p(\mathcal{R}| \mathcal{P}, \mathcal{T}_c)
    \label{equation:2}
\end{equation}

where $p(\mathcal{R}| \mathcal{P}, \mathcal{T}_c)$ and $p(\mathcal{G}_c|\mathcal{T}_c, \mathcal{P}, \mathcal{R})$  are defined as:

\begin{equation}
    p(\mathcal{R}| \mathcal{P}, \mathcal{T}_c) = \prod_{i}^{|\mathcal{R}|}  p_{LM}(r_i| \mathcal{P}, \mathcal{T}_c, r_{<i} )
    \label{equation:3}
\end{equation}

\begin{equation}
    p(\mathcal{G}_c|\mathcal{T}_c, \mathcal{P} , \mathcal{R}) = \prod_{i}^{|\mathcal{G}_c|} p_{LM}(g_i, | \mathcal{T}_c, \mathcal{P} ,\mathcal{R}, g_{<i} )
    \label{equation:4}
\end{equation}

where $r_i$ is one step of total $|\mathcal{R}|$ reasoning steps. Specifically, we decompose complex knowledge graph reasoning tasks into multiple steps, as shown in the Figure \ref{figure:rationale}. 
For the construction of knowledge graphs, the key point is to identify relations and extract their corresponding entities. 
Therefore, our designed reasoning chain includes three steps: relationship identification, entity extraction, and the final knowledge graph construction. 
Relying on our designed rationale, the model can more accurately complete the task of knowledge graph construction by utilizing the reasoning chain.

\section{Experiment}
\label{sec:experiment}

Extensive experiments are conducted to evaluate the performance of the \ours~ by answering the following research questions:
\begin{itemize}
    \item \textbf{RQ1}: How does our \ours~ perform when competing with baselines of knowledge graph construction?
    \item \textbf{RQ2}: How do key modules in our approach contribute to the overall performance?
    \item \textbf{RQ3}: How effective is the proposed \ours~ model in addressing overlapped or complex patterns? 
   \item \textbf{RQ4}: 
    How do smaller code language models affect the performance of \ours?
\end{itemize}

\subsection{Settings}

\subsubsection{{Datasets.}}
We evaluate the proposed \ours~ on three popular benchmark datasets, ADE \cite{ade}, CoNLL04 \cite{conll04} and SciERC \cite{scirec}.
\begin{itemize}
    \item  ADE \cite{ade}  is a collection of texts that focuses on the identification and extraction of adverse drug events.
    It has 1 relation (\textit{adverse effect}) and 2 entity types (\textit{disease}, \textit{drug}).
    We split the original dataset into train, valid and test set, and then select some samples from the train dataset as prompt samples under the few-shot setting.
    Limited by the time constraints of the API call, we extract 300 samples randomly chosen from the test dataset for evaluation. 

    \item The CoNLL04 \cite{conll04} dataset is a widely used benchmark for evaluating information extraction and relation extraction tasks.
    There are 4 relation (such as \textit{work for}, \textit{live in}) and 3 entity types (\textit{person}, \textit{organization}, \textit{live in}).
    We randomly select a few samples as demonstrations from the train dataset in the few-shot setting.
    
    \item SciERC \cite{scirec} represents a dataset designed for a multi-task framework, focusing on the identification and classification of entities, relations, and coreference clusters found in scientific papers.
    It consists of 7 relations (such as \textit{hyponym-of}, \textit{used-for}) and 6 entity types (such as \textit{method} and \textit{task}).
    We also randomly select several samples from the train dataset and use the whole test dataset for evaluation.
\end{itemize}

\subsubsection{Baselines for Comparison}

For \textbf{RQ1}, \textbf{RQ2} and \textbf{RQ3},
we use the large code language models (GPT-3.5 series) \textit{text-davinci-002} and \textit{text-davinci-003} as the backbone, both of which are trained on code and textual corpus.
Due to the limited use \footnote{On March 23rd, OpenAI discontinued support for the Codex (code-davinci-002) API.} of \textit{code-davinci-002 }, we only conduct ablation experiments to compare  \ours~ with different models.
\revise{
For fair comparison, we use vanilla prompt and the supervised model UIE \citep{UIE} as our main baselines.
}
\revise{
We evaluate our proposed \ours~ in both zero-shot and few-shot settings.
}
\revise{
Vanilla prompt here means the traditional prompt, which is textual and non-code structured.
}
The explanations for code models we use as backbones are shown below,
\begin{itemize}
    \item \textbf{code-davinci-002}  (also known as Codex) is a model that has been trained on both text and code corpus, and is especially good at translating natural language to code. 
    \item \textbf{text-davinci-002} is a supervised instruction-tuned model based on Codex. However, it may lose some of its code understanding abilities due to its instruction-tuning strategy.
    \item \textbf{text-davinci-003} is the most capable GPT-3 model, which is capable of handling complex instructions and generating longer-form content.
    The training corpus of \textit{text-davinci-003} also contains a large amount of code data.
    
\end{itemize}

For \textbf{RQ4},
we pre-train a smaller code language model initialized by  T5 \cite{t5} and CodeT5 \cite{CodeT5} and then evaluate the performance of our approach.
The small pre-trained model we pre-train here has only 770 million parameters, which is much less than the GPT 3.5 series models (over 10 billion parameters).
We re-structure the corpus of T-REx \cite{t-rex}, TEKGEN \cite{tekgen} and KELM \cite{tekgen} into code format (Python) and build a totally new dataset for pre-training.
Detailed statistics can be found in Table \ref{table-statistics_pt}.

\begin{table}[t]
\centering

\caption{{Dataset statistics for our smaller pre-training  models.}}

\resizebox{0.95\linewidth}{!}{
\begin{tabular}{llllll}
\toprule
\textbf{Data Source} & {\textbf{Sentences}} & {\textbf{Triples}} & {\textbf{Relations}} & {\textbf{Descriptions}}\\
\midrule
T-REx~\cite{t-rex} & 838,821  &  4,378,318 & 10,496 & \makecell[l]{Large-scale corpus having alignments between \\ Wikidata triples and DBpedia abstracts} \\
TEKGEN~\cite{tekgen} & 6,310,061 & 10,609,697 &  1,013 & \makecell[l]{Large-scale corpus constructed by distant super-\\-vision, which was previously used for training\\ data-to-text generation models}\\
KELM~\cite{tekgen} & 15,616,551  & 35,117,608  &  59,904 & \makecell[l]{Synthetic corpus containing the entire Wikidata\\ with text sentences}\\
\midrule
Overall ~ & 22,765,433  & 50,105,623 & 71,413 & Corpus in code format \\ 
\bottomrule
\end{tabular}
}

\label{table-statistics_pt}
\end{table}

\subsubsection{{Evaluation Protocols}}
We calculate the relation strict micro F1 score for the knowledge graph construction task.
It is important to note that a triple is considered correct only if the head and tail entities and the relation in the triplet completely match. 
Although our model can predict the corresponding types of entities, these types are not included when calculating the F1 score.
\revise{
For all experimental setups, we run three times and use the average results in the end to ensure fairness.
}

\subsubsection{{Parameters Settings}}
For \textbf{RQ1}, \textbf{RQ2} and \textbf{RQ3}, we utilize the API interface provided by OpenAI and  set the \textit{temperature} to 0.5 and  \textit{max\_tokens} to either 256 or 512.
Meanwhile, we stop the code generation whenever any of the special patterns such as """, class or \#.
To solve \textbf{RQ4}, we pre-train the small code language model by utilizing Pytorch 1.13 with deepspeed\footnote{\url{https://www.deepspeed.ai}} package and  conduct experiments on 4 NVIDIA Tesla V100 GPUs for three days.
In the fine-tuning stage of small language models, all optimization is performed by using the AdamW optimizer. 

\subsection{Results and Analysis}
\label{sec:results}

\subsubsection{Main Results}

\begin{table*}[t]
\centering
\caption{
\revise{
Performance of {\ours} in zero-shot and few-shot settings (relation strict micro F1).
}
The \textbf{bold} numbers denote the best results and the  \underline{underlined} ones are the second-best performance.
We also report score improvement  results ($\uparrow$) over the same backbone model.
For the zero-shot setting, we replace {1-shot} relations with their synonyms.
For the few-shot setting, we choose {3-shot} relation setting as demonstrations.
}
\begin{tabular}{llccc}
\toprule
& \multirow{2}{*}{\textbf{Comparable SOTA} } & \multicolumn{3}{c}{\textbf{Dataset}} \\ \cline{3-5}
 &  & \textbf{ADE}  & \textbf{CONLL04}  & \textbf{SciERC}  \\ 
\midrule
\multirow{1}{*}{{\textbf{Zero-Shot}} } 
& \revise{UIE \cite{UIE} } &  24.3  & 16.1   &  10.3 \\
\cline{2-5}
& Vanilla Prompt (text-davinci-002) & {41.2} & 18.4 & 12.2\\ 
& Vanilla Prompt (text-davinci-003) & 41.7  & {30.5} & \underline{18.1} \\
\cline{2-5}
& CodeKGC (text-davinci-002) & \underline{42.5}~($\uparrow$1.3) & \underline{35.8}~($\uparrow$17.4) & 15.0~($\uparrow$2.8) \\ 
& CodeKGC (text-davinci-003) & \textbf{43.7}~($\uparrow$2.0)  & \textbf{41.6}~($\uparrow$11.1) & \textbf{19.5}~($\uparrow$1.4)\\ 
\midrule
\multirow{2}{*}{\textbf{Few-Shot} }
& \revise{UIE \cite{UIE} } &  50.3 &  39.0  & \underline{19.2} \\ 
\cline{2-5}
& Vanilla Prompt (text-davinci-002) & 45.7 & 28.2 & 14.1 \\ 
& Vanilla Prompt (text-davinci-003) & {58.8} & \underline{43.2} & {18.8} \\ 
\cline{2-5}
& CodeKGC (text-davinci-002) & \underline{61.5}~($\uparrow$15.8) & 42.7~($\uparrow$14.5) & 18.5~($\uparrow$4.4) \\ 
& CodeKGC (text-davinci-003) & \textbf{64.2}~($\uparrow$5.4) & \textbf{49.6}~($\uparrow$6.4) & \textbf{24.7}~($\uparrow$5.9) \\ 
\bottomrule
\end{tabular}
\label{results-main}
\end{table*}

Table \ref{results-main} shows the performance of \ours~ in zero-shot and few-shot settings.
In the zero-shot setting, we select one sample for each relation and replace the relations with their synonyms.
The experimental results demonstrate that our proposed \ours~ outperforms \revise{traditional supervised method UIE}  and conventional text prompts on all three datasets.
\revise{We note that our approach exhibits substantial score improvements especially in CONLL04.}
We also note that the performance of {\ours}  on the ADE dataset is limited. 
This is because the ADE dataset is a medical corpus with a large number of acronyms, which makes extraction more difficult. Therefore, in the zero-shot setting, the performance of {\ours} is similar to that of Vanilla Prompt.

In the few-shot setting, we choose 3-shot relation setting as demonstrations.
\revise{We find our proposed approach obtains excellent performance compared to UIE.}
\revise{Compared to Vanillar Prompt}, \ours~ proves more effective when employing \textit{text-davinci-003}.
Specifically, \ours~ achieves \revise{$\uparrow$5.4, $\uparrow$6.4 and $\uparrow$5.9}  score improvement on ADE, CONLL04 and SciERC datasets, which might be due to the fact that \textit{text-davinci-003} utilizes higher-quality code data \revise{in the pre-training stage}.
Although \textit{text-davinci-003} itself performs better than \textit{text-davinci-002}, {\ours} with \textit{text-davinci-002} remains competitive to Vanilla Prompt with \textit{text-davinci-003}.
{\ours} even with \textit{text-davinci-002} performs better than Vanilla Prompt with \textit{text-davinci-003} on the ADE dataset.
It demonstrates the effectiveness of our designed code-formatted prompts in enabling the language model to efficiently extract the requisite entities and relations for knowledge graph construction from the text.

\begin{figure}
\centering
\includegraphics[width=0.63\textwidth]{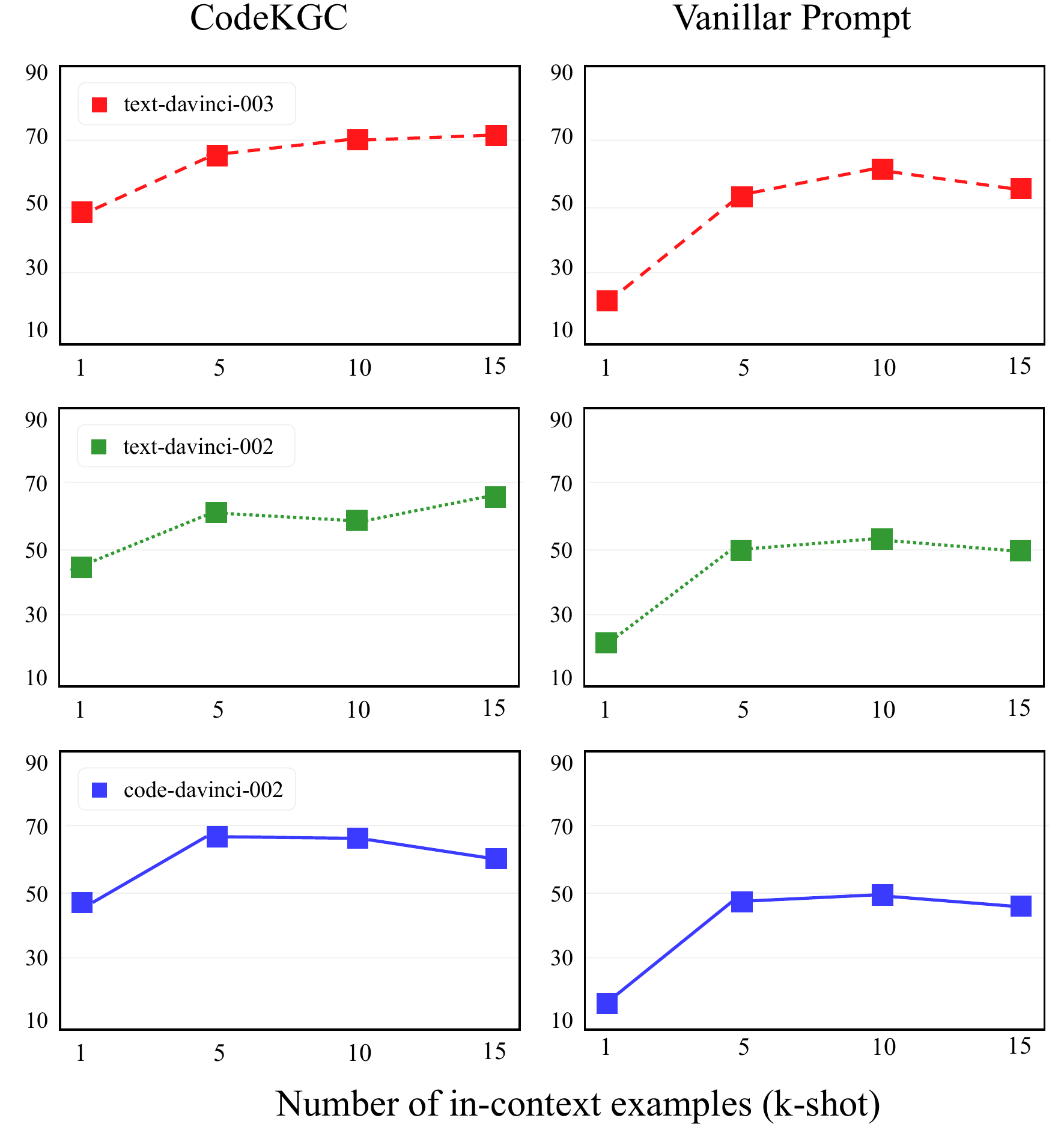}
\caption{
\revise{
Few-shot performance comparison in ADE dataset between code-based prompt (\ours) and text-based prompt (\textbf{Vanilla Prompt}) on different models (relation strict micro F1).
}
}
\label{results-main2}
\end{figure}

To further investigate the impact of the number of in-context samples, we evaluate the performance of {\ours} and Vanilla Prompt on the ADE dataset using different settings of \{1, 5, 10, 15\}.
In addition to \textit{text-davinci-002} and \textit{text-davinci-003}, we also select another code model (Codex) as the backbone to further investigate the impact of the Code large model.
From Figure \ref{results-main2}, we notice that \ours~ obtains better few-shot performance than Vanilla Prompt.
We observe that, as the number of in-context samples increases, our {\ours} consistently outperforms the baseline in all models (\textit{text-davinci-002}, \textit{text-davinci-003} and \textit{code-davinci-002}),
indicating that the language of code indeed has the better ability to model structure and address the \textbf{RQ1}.

Another significant finding in Figure \ref{results-main2} is that when the same {\ours} is applied with \textit{text-davinci-002} and \textit{code-davinci-002} as the backbone, the performance of \textit{code-davinci-002} is better than that of \textit{text-davinci-002}.
It might be because \textit{code-davinci-002} has stronger performance on code-related tasks such as code generation and code completion tasks.
Therefore, \textit{code-davinci-002} shows more advantages in information extraction tasks.

\subsubsection{Results of Prompt Variations of {\ours}}

\begin{figure}[h]
\centering
\includegraphics[width=0.5\textwidth]{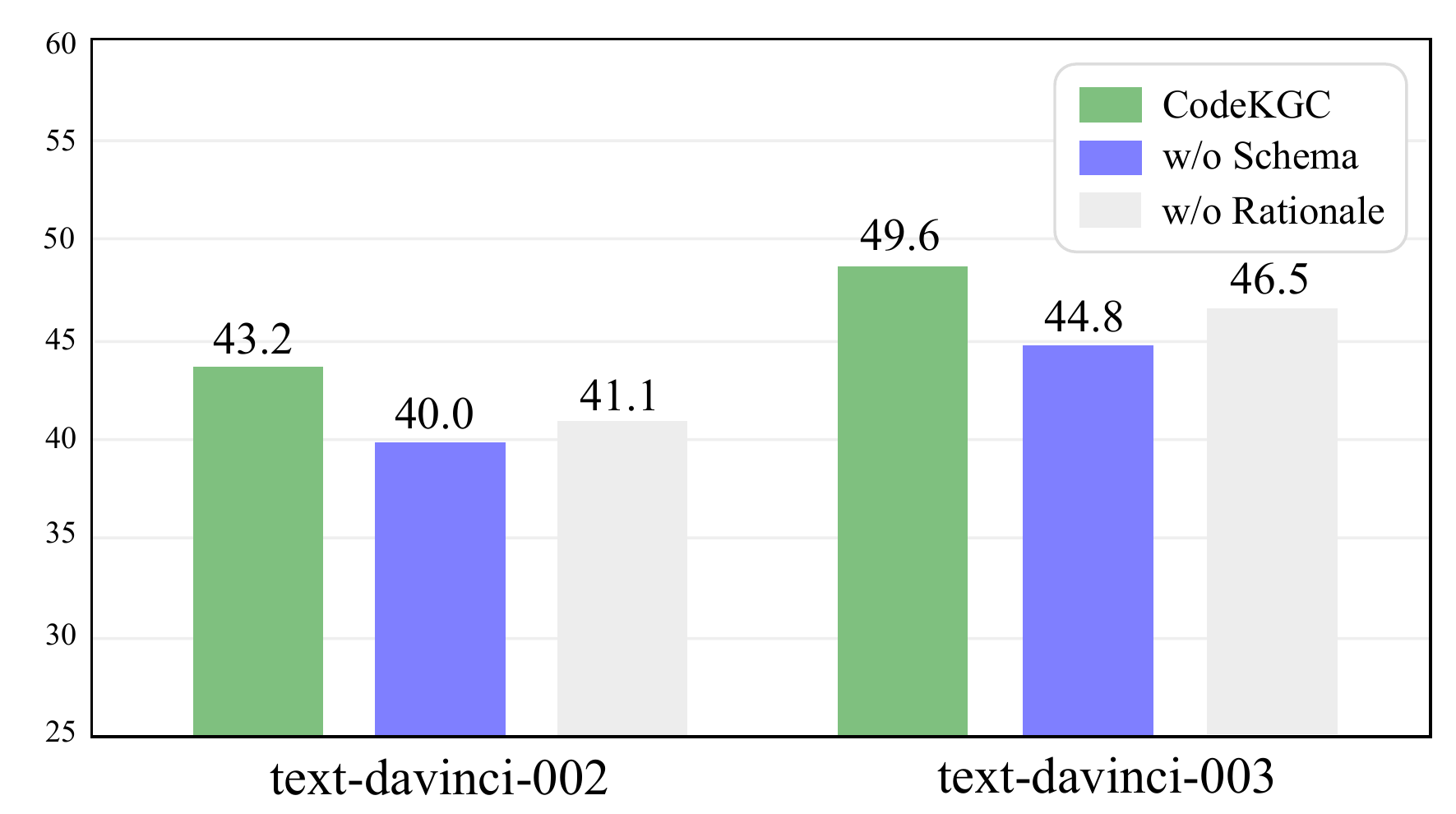}
\caption{
\revise{
Impacts of prompt variation in the CoNLL04 dataset. 
\textit{w/o} is without schema-aware prompt and rationales. 
Our designed schema-aware prompt and rationale-enhanced generation obviously enhance the performance of our proposed approach.
}
}
\label{figure-rationale}
\end{figure}

To address the \textbf{Q2}, we conduct experiments to investigate the impacts of key module variations of our proposed {\ours}.

\revise{
\paragraph{\textbf{Impacts of schema-aware prompt}}
To study the impacts of schema-aware prompt, we choose the dataset CoNLL04, which consist of multiple relations and entities.
In Figure \ref{figure-rationale}, we compare the result when {\ours} is prompt with and without schema-aware prompt.
We find that our proposed schema-aware method can effectively encode the structural feature.
By integrating the schema information in the knowledge graph, {\ours} can efficiently leverage the powerful encoding abilities of code language models.
}


\paragraph{\textbf{Impacts of rationale-enhanced generation}}
The results from Figure \ref{figure-rationale} on the CoNLL04 dataset show that rationale contributes to improving the performance of {\ours} on \textit{text-davinci-002} and \textit{text-davinci-003}.
Incorporating rationales in knowledge graph construction tasks is beneficial because rationales assist in guiding the model's focus on specific aspects of the input text during generation, leading to more effective learning and improved understanding of complex relationships in the graph.
It should be noted that we also find limitations with the rationale during the experiment. 
When the sample size is too small, lengthy prompts can have a negative impact on the model.

\subsubsection{Results of Handling Hard Examples}

\begin{figure}[t!]
\centering
\includegraphics[width=0.5\textwidth]{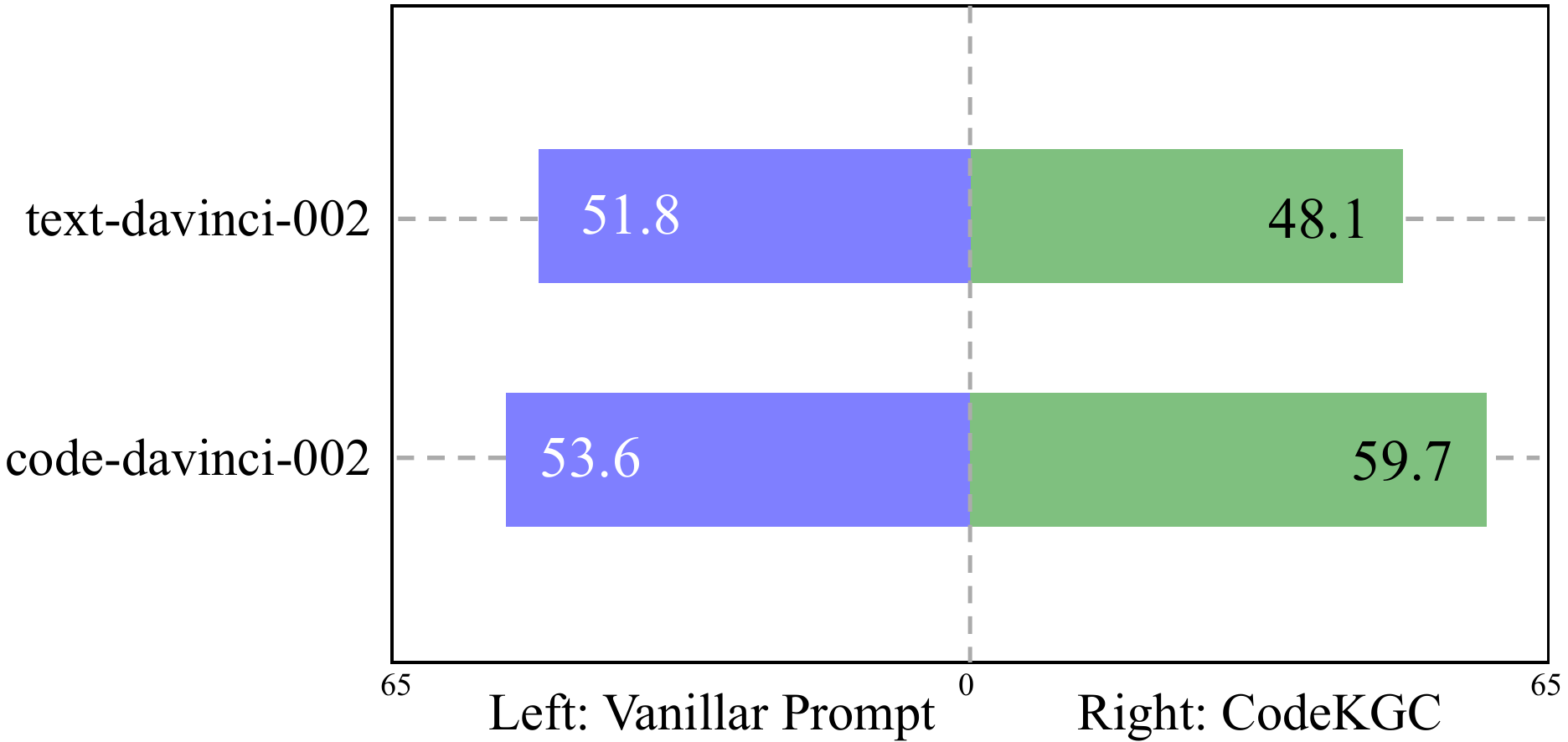}
\caption{
\revise{
Few-shot prediction of \ours~ and \textbf{Vanilla Prompt} in hard examples (relation strict micro F1). 
}
\ours~ with \textit{code-davinci-002 } correctly predicts the overlapping triples, while Vanilla Prompt fails.
}
\label{figure-hard_example_1}
\end{figure}

\begin{figure}[t!]
\centering
\includegraphics[width=0.55\textwidth]{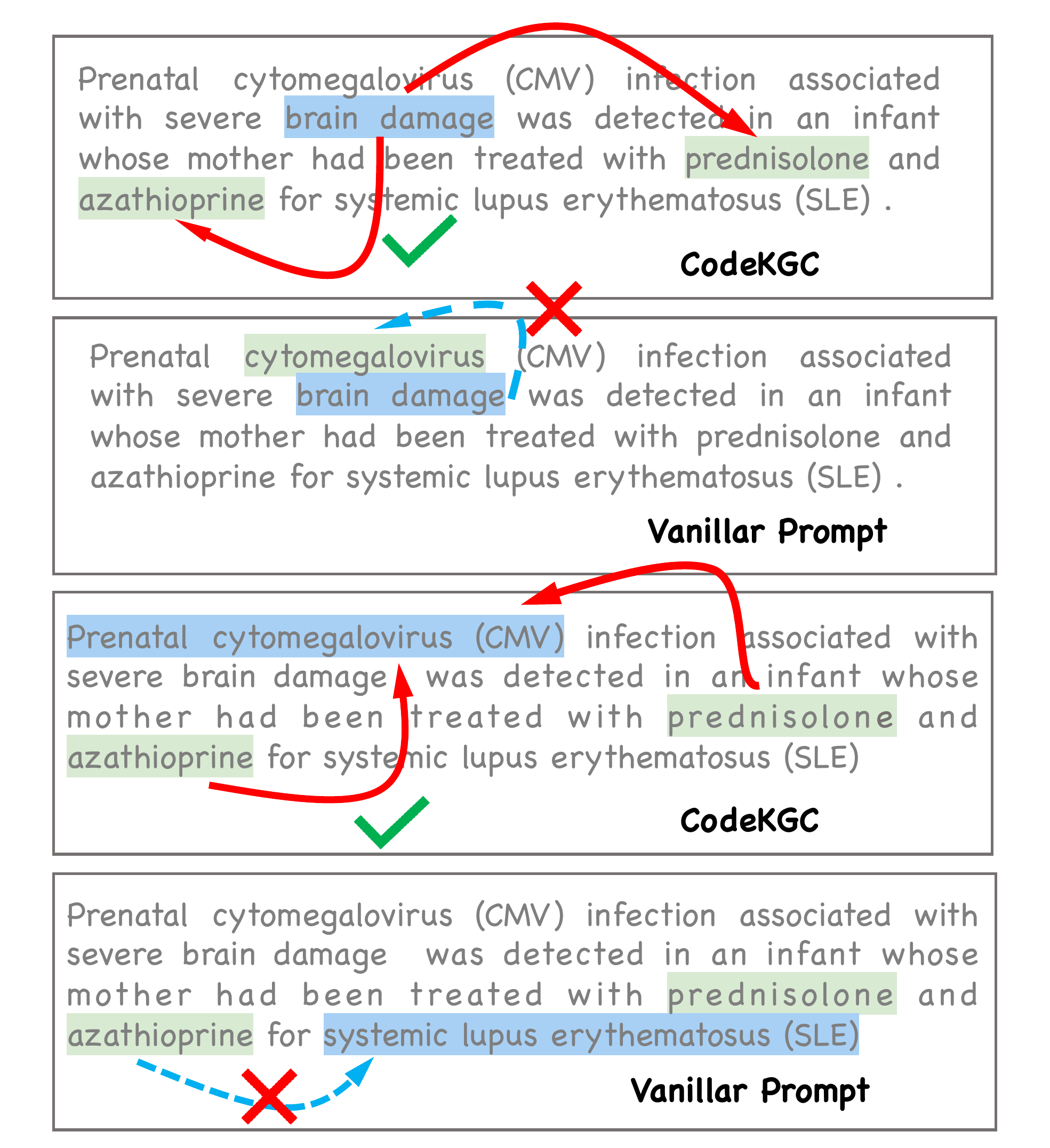}
\caption{
Cases of \ours~ and \textbf{Vanilla Prompt} in hard examples. 
\ours~ with \textit{code-davinci-002 } correctly predicts the overlapping and long-range triples, while Vanilla prompt fails.
}
\label{figure-hard_example_2}
\end{figure}


\revise{
To investigate overlapping issues, we focus on extracting multiple intersecting or overlapping triples from the same text.
We sample the overlapping triples, termed as 'hard samples', from the ADE dataset and then evaluate the performance of {\ours} and Vanilla Prompt using \textit{text-davinci-002} and \textit{code-davinci-002} as backbones.
}
\revise{
In Figure \ref{figure-hard_example_1}, we compare the results in few-shot setting, which are evalauted in the hard samples.
}
We observe that {\ours} with \textit{code-davinci-002} achieves better results compared to Vanilla Prompt with \textit{text-davinci-002}.
It shows that {\ours} can enhance the performance of generating overlapping triples.
\revise{
However, we also notice that the performance of Vanillar Prompt (\textit{text-davinci-002}) is significantly better than that of CodeKGC(\textit{text-davinci-002}).
Although \textit{text-davinci-002} is also a hybrid code model, CodeKGC heavily relies on a more powerful code pre-trained model as its foundation when solving  complex structures. It is also a limitation of our current approach.
}

\revise{
As depicted in Figure \ref{figure-hard_example_2}, we select some cases from the prediction result and  find that {\ours} is powerful in handling the cases of long-distance dependencies and overlapping issues.
}
\revise{
When both two overlapped triples (the entity \textit{brain damage} and \textit{CMV}) are nested,  our approach can accurately extract the relations and entities, which demonstrates superiority in addresses \textbf{Q3}.
}
It is also worth noting that vanilla natural language generation models with flattened strings tend to push neighboring nodes farther apart, while code language preserves structural information, thereby improving the performance in handling overlapping cases.

\subsubsection{Analysis of Smaller Code Language Models}


\revise{
To address \textbf{Q4}, we do not directly choose to use existing large pre-trained code models with a huge number of parameters (such as GPT-3.5 serious models). 
Instead, we pre-train a model with fewer parameters on our own.
Specifically, we build a totally new pre-training dataset by re-structuring structural knowledge into code corpus and the triples in the source knowledge base (Table \ref{table-statistics_pt}) are converted into code format.
Then we use T5 \cite{t5} and CodeT5 \cite{CodeT5} as backbone and pre-train the new smaller code language model on our build code dataset.
Finally, we use four pre-trained models (pre-trained T5 and CodeT5, not pre-trained T5 and CodeT5) to fine-tune our {\ours} on the ADE dataset and compare their performance.
}

\begin{figure}
\centering
\includegraphics[width=0.43\textwidth]{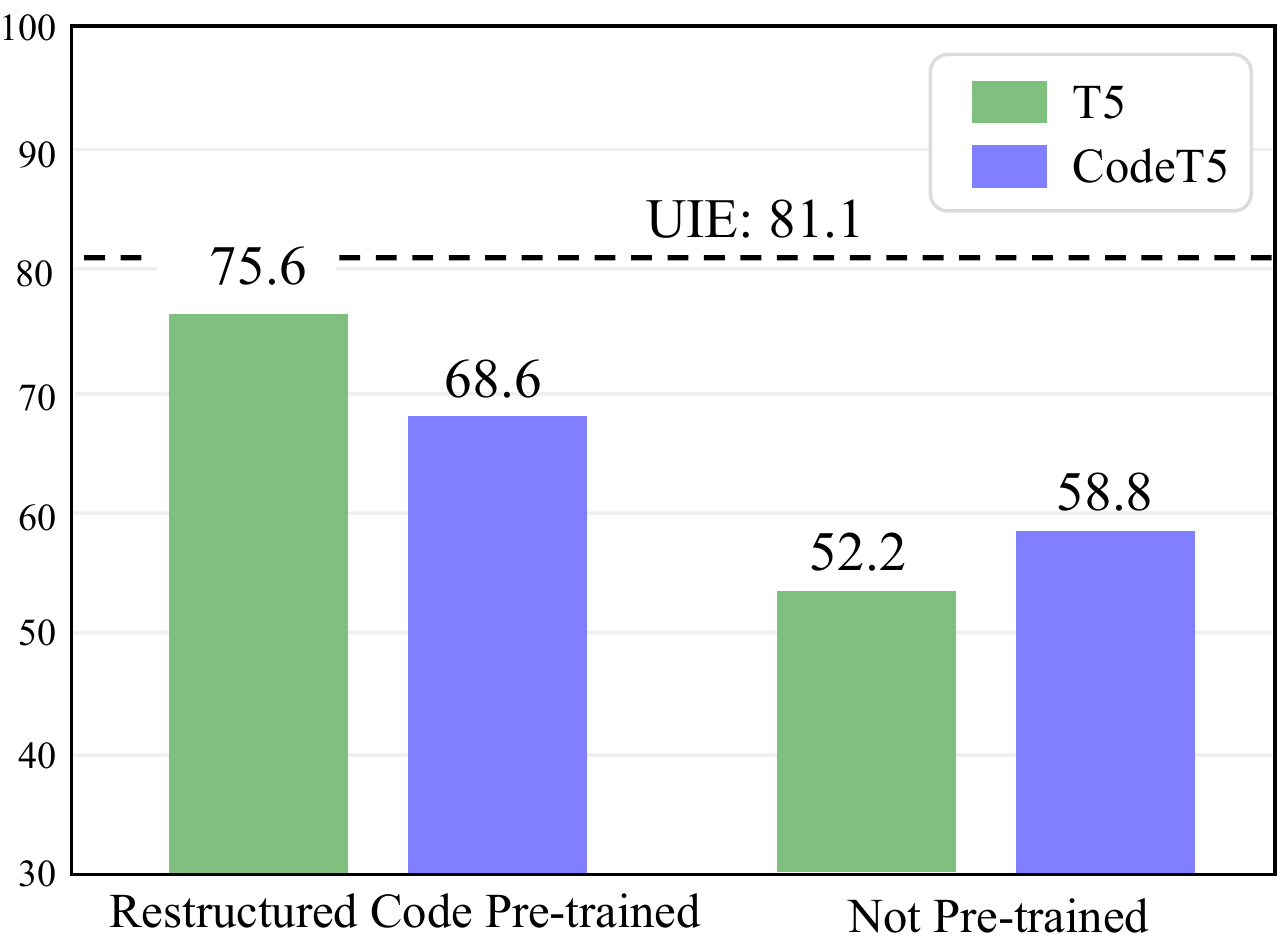}
\caption{\ours~ performance in ADE dataset with small code LMs and (relation strict micro F1).}
\label{figure-lm_small}
\end{figure}

From Figure \ref{figure-lm_small}, when both CodeT5 and T5 are not further pre-trained, CodeT5 performs significantly better which is because CodeT5 inherently understands structured code knowledge.
However, we notice that pre-training with re-structured T5  performs better than re-structured CodeT5, indicating that it is crucial for knowledge graph construction models to understand textual semantics.
Furthermore, we also test the performance of the SOTA model UIE \cite{UIE} and {\ours} performs slightly worse compared to UIE.
We argue that for knowledge graph construction as code generation, the model should have the ability of semantic understanding and code generation.
\revise{
We can't have our cake and eat it too due to the limited parameter space.
}
Such powerful ability may be emerged by large language models \cite{DBLP:journals/corr/abs-2206-07682}. 

\section{Discussion}
\label{sec:discussion}

\paragraph{\textbf{Data Format for Structure Prediction}}
\revise{The data format is essential for the knowledge graph construction task,  which is fundamentally the structure prediction problem.}
Many existing methods either process natural language in a flattened way or exploit structural information through constrained decoding techniques.
However, such explicit modeling of structural information does not effectively address the knowledge graph construction problem.
We have noticed that, in addition to natural language, code language inherently carries structural information. 
Although some work (like UIE \cite{UIE} and DeepStruct \cite{DeepStruct}) has attempted to handle structural information by designing specific structural languages, code language is more prevalent and universal in practice.
Lastly, we also observe that with the success of LLMs like ChatGPT, a wealth of instruction tuning data is available.
Instruction tuning has proven that interactive natural language is a more effective data format and can enable large language models to better understand structural information.

\paragraph{\textbf{Code Language vs. Natural Language} }
Code language and natural language are two distinct forms of language. 
Code language (programming language) serves as a means of communication between humans and machines, characterized by strict grammar and syntax rules. 
Natural languages, on the other hand, are primarily human-readable languages, used to express a wide range of ideas, emotions, and intentions. 
Both types of language can convey semantics, but the semi-structured nature of code languages allows them not only to include elements of natural language but also to serve as a bridge for better modeling of structured information. 
Furthermore, as programmable and executable languages, code languages inherently possess greater logical coherence, providing a natural advantage for certain reasoning tasks. 
Therefore, the research for code languages holds academic significance and scientific value.

\section{Conclusion and Future work}
\label{sec:conclusion}

In this paper, we have presented a novel method {\ours} that leverages the code language model for generative knowledge graph construction.
By converting natural language into code formats, our approach effectively encodes the schema structure of entities and relationships in the knowledge graph. 
We also propose rationale-enhanced generation method, which contributes to the decoupling of knowledge within the language model.
Extensive experiment results reveal that the proposed model can obtain better performance for the knowledge graph construction task.

In future work, we plan to further explore the capabilities of our {\ours} method in several directions:
1) Integration of more advanced Code-LLMs:
As the field of language models for code continues to evolve, we will incorporate newer and more powerful Code-LLMs into our framework, which might lead to further improvements in knowledge graph construction performance.
2) Application to additional knowledge graph tasks: 
Our approach can be extended to more knowledge graph-related tasks, such as knowledge graph completion or complex reasoning tasks. 
Exploring the applicability and effectiveness of our method will help to further validate its versatility.
3) Optimization of code prompt generation: 
The current code prompts we design a can be further refined to better capture the nuances of the underlying entity and relation structures.
Leveraging techniques from prompt engineering or automatic prompt generation can also help to improve the quality and effectiveness of the code prompts.
\revise{
4) Impact of code data to  LLMs:
we will continue to investigate a more critical question, exploring what form of code data truly endows LLMs with powerful structural prediction and reasoning capabilities.
}

\section*{AUTHOR CONTRIBUTION}
{Zhen Bi}: Conceptualization, Methodology, Software, Writing–original draft. 
{Jing Chen}: Conceptualization, Methodology, Software, Writing–review \& editing. 
{Yinuo Jiang}: Methodology, Software.
{Feiyu Xiong}: Conceptualization, Writing–review \& editing.
{Wei Guo}: Conceptualization, Investigation.
{Huajun Chen}: Supervision, Project administration.
{Ningyu Zhang}: Supervision, Project administration, Funding acquisition;

\begin{acks}
We would like to express gratitude to the anonymous reviewers for their kind comments. This work was supported by the National Natural Science Foundation of China (No. 62206246), the Fundamental Research Funds for the Central Universities (226-2023-00138), Zhejiang Provincial Natural Science Foundation of China (No. LGG22F030011), Ningbo Natural Science Foundation (2021J190), CAAI-Huawei MindSpore Open Fund, Yongjiang Talent Introduction Programme (2021A-156-G), CCF-Baidu Open Fund, and Information Technology Center and State Key Lab of CAD\&CG, Zhejiang University. 
\end{acks}


\bibliographystyle{ACM-Reference-Format}
\bibliography{sample-base}

\end{document}